\def\year{2016}\relax
\documentclass[letterpaper]{article}
\usepackage{aaai16}
\usepackage{times}
\usepackage{helvet}
\usepackage{courier}
\usepackage{tabularx}
\usepackage{ragged2e}  
\usepackage{graphicx}

\usepackage{tikz-dependency}

\newcolumntype{Y}{>{\RaggedRight\arraybackslash}X} 
\begin{document}
%
\title{Tweet Acts: A Speech Act Classifier for Twitter{}}

\author{Soroush Vosoughi, Deb Roy\\
Massachusetts Institute of Technology\\ Cambridge, MA 02139\\ \tt{soroush@mit.edu, dkroy@media.mit.edu}}

\maketitle
\begin{abstract}
\begin{quote}
Speech acts are a way to conceptualize speech as action. This holds true for communication on any platform, including social media platforms such as Twitter. In this paper, we explored speech act recognition on Twitter by treating it as a multi-class classification problem. We created a taxonomy of six speech acts for Twitter and proposed a set of semantic and syntactic features. We trained and tested a logistic regression classifier using a data set of manually labelled tweets. Our method achieved a state-of-the-art performance with an average F1 score of more than $0.70$. We also explored classifiers with three different granularities (Twitter-wide, type-specific and topic-specific) in order to find the right balance between generalization and overfitting for our task. 

\end{quote}
\end{abstract}

\section{Introduction}
In recent years, the micro-blogging platform Twitter has become a major social media platform with hundreds of millions of users. People turn to Twitter for a variety of purposes, from everyday chatter to reading about breaking news. The volume plus the public nature of Twitter (less than 10\% of Twitter accounts are private \cite{techcrunch2009}) have made Twitter a great source of data for social and behavioural studies. These studies often require an understanding of what people are tweeting about. Though this can be coded manually, in order to take advantage of the volume of tweets available automatic analytic methods have to be used. 
There has been extensive work done on computational methods for analysing the linguistic content of tweets. 
However, there has been very little work done on classifying the pragmatics of tweets. Pragmatics looks beyond the literal meaning of an utterance and considers how context and intention contribute to meaning. A major element of pragmatics is the intended communicative act of an utterance, or what the utterance was meant to achieve. 

It is essential to study pragmatics in any linguistic system because at the core of linguistic analysis is studying what language is used for or what we do with language. Linguistic communication and meaning can not truly be studied without studying pragmatics. Proposed by Austin \cite{john1962austin} and refined by Searle \cite{searle1969speech}, speech act theory can be used to study pragmatics. Amongst other things, the theory provides a formalized taxonomy \cite{searle1976taxonomy} of a set of communicative acts, more commonly known as speech acts. 

There has been extensive research done on speech act (also known as dialogue act) classification in computational linguistics, e.g., \cite{stolcke2000dialogue}. Unfortunately, these methods do not map well to Twitter, given the noisy and unconventional nature of the language used on the platform. In this work, we created a supervised speech act classifier for Twitter, using a manually annotated dataset of a few thousand tweets, in order to be better understand the meaning and intention behind tweets and uncover the rich interactions between the users of Twitter. Knowing the speech acts behind a tweet can help improve analysis of tweets and give us a better understanding of the state of mind of the users. Moreover, ws we have shown in our previous works \cite{vosoughi_rd_2015,vosoughi2015automatic}, speech act classification is essential for detection of rumors on Twitter. Finally, knowing the distribution of speech acts of tweets about a particular topic can reveal a lot about the general attitude of users about that topic (e.g., are they confused and are asking a lot of questions? Are they outraged and demanding action? Etc).  

\section{Problem Statement}
Speech act recognition is a multi-class classification problem. As with any other \emph{supervised} classification problem, a large labelled dataset is needed. In order to create such a dataset we first created a taxonomy of speech acts for Twitter by identifying and defining a set of commonly occurring speech acts. Next, we manually annotated a large collection of tweets using our taxonomy. Our primary task was to use the expertly annotated dataset to analyse and select various syntactic and semantic features derived from tweets that are predictive of their corresponding speech acts. Using our labelled dataset and robust features we trained standard, off-the-shelf classifiers (such as SVMs, Naive Bayes, etc) for our speech act recognition task.

Using Searle's speech act taxonomy \cite{searle1976taxonomy}, we established a list of six speech act categories that are commonly seen on Twitter: \emph{Assertion}, \emph{Recommendation} \emph{Expression}, \emph{Question}, \emph{Request}, and \emph{Miscellaneous}. Table \ref{table:speech-acts} shows an example tweet for each of these categories.

\begin{table}[h]
\centering
\begin{tabularx}{\columnwidth}{@{} l| Y @{}}
\hline
Act & Example Tweet  \\
 \hline
Asr & authorities say that the 2 boston bomb suspects are brothers are legal permanent residents of chechen origin - @nbcnews\\ 
Rec & If you follow this man for updates and his opinions on \#Ferguson  I recommend you unfollow him immediately.  \\ 
Exp & Mila Kunis and Ashton Kutcher are so adorable \\ 
Que &  Anybody hear if @gehrig38 is well enough to attend tonight? \#redsox \\ 
Req & rt @craigyh999: 3 days until i run the london marathon in aid of the childrens hopsice @sschospices . please please sponsor me here \\ 
Mis & We'll continue to post information from \#Ferguson throughout the day on our live-blog
\end{tabularx}
\caption{Example tweets for each speech act type.}

\label{table:speech-acts}

\end{table}






\section{Data Collection and Datasets}
Given the diversity of topics talked about on Twitter, we wanted to explore topic and type dependent speech act classifiers. We used Zhao et al.'s \cite{zhao2011empirical} definitions for topic and type. A \emph{topic} is a subject discussed in one or more tweets (e.g., Boston Marathon bombings, Red Sox, etc). The \emph{type} characterizes the nature of the topic, these are: 
\emph{Entity-oriented}, \emph{Event-oriented topics}, and \emph{Long-standing topics} (topics about subjects that are commonly discussed).

\begin{figure}[!htbp]
\centering
\includegraphics[width=1.\columnwidth]{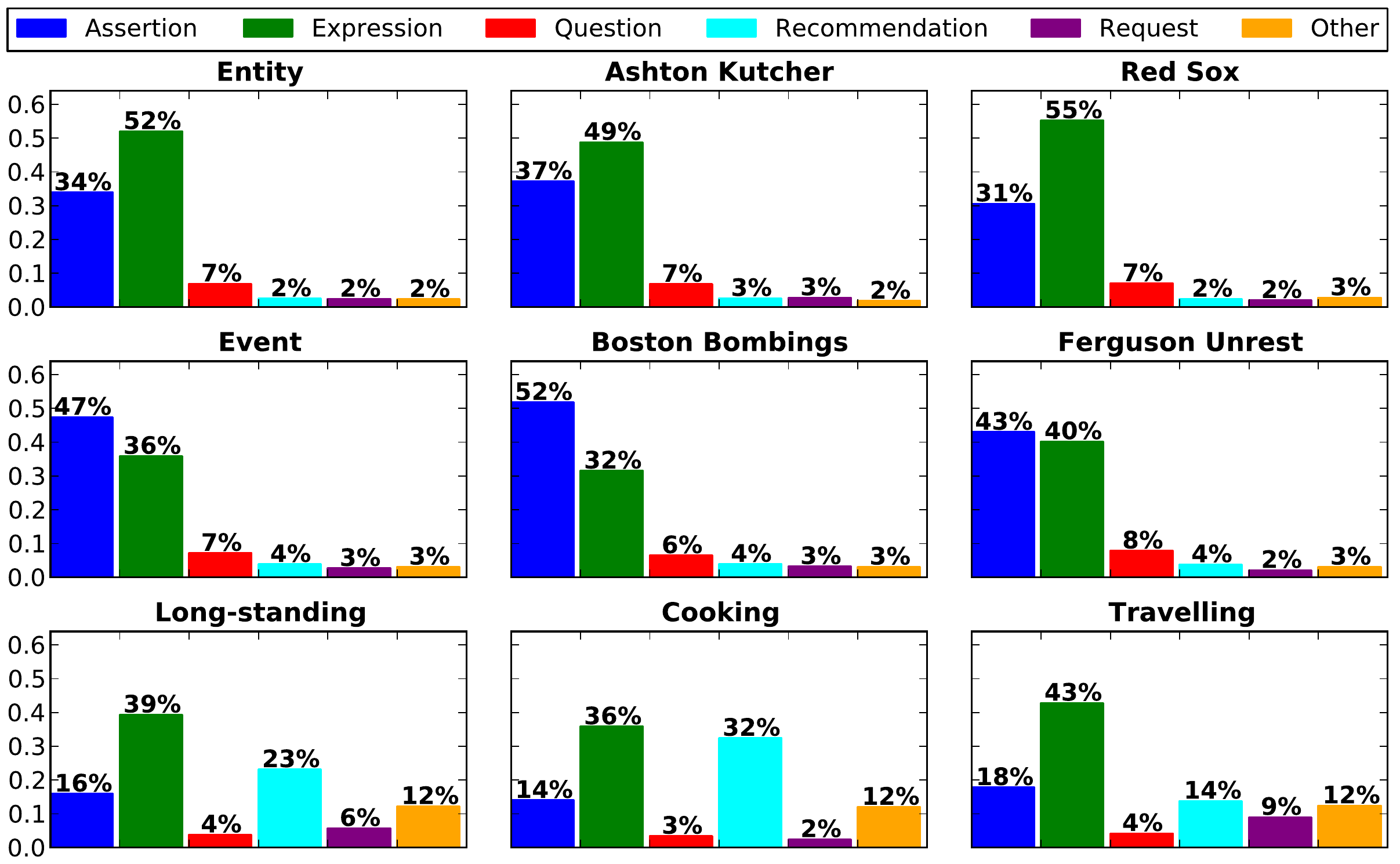}   
\caption{Distribution of speech acts for all six topics and three types.}
\label{fig:SAdivisions}
\end{figure}

We selected two topics for each of the three topic types described in the last section for a total of six topics (see Figure \ref{fig:SAdivisions} for list of topics). We collected a few thousand tweets from the Twitter public API for each of these topics using topic-specific queries (e.g., \#fergusonriots, \#redsox, etc). We then asked three undergraduate annotators to independently annotate each of the tweets with one of the speech act categories described earlier. The \emph{kappa} for the annotators was $0.78$. For training, we used the label that the majority of annotators agreed upon (7,563 total tweets). 


			  



The distribution of speech acts for each of the six topics and three types is shown in Figure \ref{fig:SAdivisions}. There is much greater similarity between the distribution of speech acts of topics of the same type (e.g, Ashton Kutcher and Red Sox) compared to topics of different types. Though each topic type seems to have its own distinct distribution, \emph{Entity} and \emph{Event} types have much closer resemblance to each other than \emph{Long-standing}. Assertions and expressions dominate in \emph{Entity} and \emph{Event} types with questions beings a distant third, while in \emph{Long-standing}, recommendations are much more dominant with assertions being less so. This agrees with Zhao et al.'s \cite{zhao2011empirical} findings that tweets about \emph{Long-standings} topics tend to be more opinionated which would result in more recommendations and expressions and fewer assertions.

The great variance across types and the small variance within types suggests that a type-specific classifier might be the correct granularity for Twitter speech act classification (with topic-specific being too narrow and Twitter-wide being too general). We will explore this in greater detail in the next sections of this paper.

\section{Features}
We studied many features before settling on the features below. Our features can be divided into two general categories: \emph{Semantic} and \emph{Syntactic}. Some of these features were motivated by various works on speech act classification, while others are novel features. Overall we selected $3313$ binary features, composed of $1647$ semantic and $1666$ syntactic features.  

\subsection{Semantic Features}
\emph{\textbf{Opinion Words: }}We used the "Harvard General Inquirer" lexicon \cite{stone1968general}, which is a dataset used commonly in sentiment classification tasks, to identify $2442$ strong, negative and positive opinion words (such as \emph{robust}, \emph{terrible}, \emph{untrustworthy}, etc). The intuition here is that these opinion words tend to signal certain speech acts such as expressions and recommendations. One binary feature indicates whether any of these words appear in a tweet. 

\noindent
\newline
\emph{\textbf{Vulgar Words: }}
Similar to opinion words, vulgar words can either signal great emotions or an informality mostly seen in expressions than any other kind of speech act (least seen in assertions). We used an online collection of vulgar words\footnote{http://www.noswearing.com/dictionary} to collect a total of $349$ vulgar words. A binary feature indicates the appearance or lack thereof of any of these words.

\noindent
\newline
\emph{\textbf{Emoticons: }}
Emoticons have become ubiquitous in online communication and so cannot be ignored. Like vulgar words, emoticons can also signal emotions or informality. We used an online collection of text-based emoticons\footnote{http://pc.net/emoticons/} to collect a total of $362$ emoticons. A binary feature indicates the appearance or lack thereof of any of these emoticons.

\noindent
\newline
\emph{\textbf{Speech Act Verbs: }}There are certain verbs (such as \emph{ask}, \emph{demand}, \emph{promise}, \emph{report}, etc) that typically signal certain speech acts. Wierzbicka \cite{wierzbicka1987english} has compiled a total of $229$ English speech act verbs divided into $37$ groups. 
Since this is a collection of verbs, it is crucially important to only consider the verbs in a tweet and not any other word class (since some of these words can appear in multiple part-of-speech categories). In order to do this, we used Owoputi et al.'s \cite{owoputi2013improved} Twitter part-of-speech tagger to identify all the verbs in a tweet, which were then stemmed using \emph{Porter Stemming} \cite{porter1980algorithm}. The stemmed verbs were then compared to the $229$ speech act verbs (which were also stemmed using Porter Stemming). Thus, we have $229$ binary features coding the appearance or lack thereof of each of these verbs.

\noindent
\newline
\emph{\textbf{N-grams: }}In addition to the verbs mentioned, there are certain phrases and non-verb words that can signal certain speech acts. For example, the phrase \emph{"I think"} signals an expression, the phrase \emph{"could you please"} signals a request and the phrase \emph{"is it true"} signals a question. Similarly, the non-verb word \emph{"should"} can signal a recommendation and \emph{"why"} can signal a question.

These words and phrases are called n-grams (an n-gram is a contiguous sequence of n words). Given the relatively short sentences on Twitter, we decided to only consider unigram, bigram and trigram phrases. We generated a list of all of the unigrams, bigrams and trigrams that appear at least five times in our tweets for a total of 6,738 n-grams. From that list we selected a total of 1,415 n-grams that were most predictive of the speech act of their corresponding tweets but did not contain topic-specific terms (such as \emph{Boston}, \emph{Red Sox}, etc). There is a binary feature for each of these sub-trees indicating their appearance.

\subsection{Syntactic Features}
\emph{\textbf{Punctuations: }}Certain punctuations can be predictive of the speech act in a tweet. Specifically, the punctuation \emph{?} can signal a question or request while \emph{!} can signal an expression or recommendation. We have two binary features indicating the appearance or lack thereof of these symbols.

\noindent
\newline
\emph{\textbf{Twitter-specific Characters: }}There are certain Twitter-specific characters that can signal speech acts. These characters are \emph{\#}, \emph{@}, and \emph{RT}.The position of these characters is also important to consider since Twitter-specific characters used in the initial position of a tweet is more predictive than in other positions. Therefore, we have three additional binary features indicating whether these symbols appear in the initial position.  


\noindent
\newline
\emph{\textbf{Abbreviations: }}
Abbreviations are seen with great frequency in online communication. The use of abbreviations (such as \emph{b4} for \emph{before}, \emph{jk} for \emph{just kidding} and \emph{irl} for \emph{in real life}) can signal informal speech which in turn can signal certain speech acts such as expression. We collected $944$ such abbreviations from an online dictionary\footnote{http://www.netlingo.com/category/acronyms.php} and Crystal's book on language used on the internet \cite{crystal2006language}. We have a binary future indicating the presence of any of the $944$ abbreviations.

\begin{figure}[!htbp]
\centering
\includegraphics[width=\columnwidth]{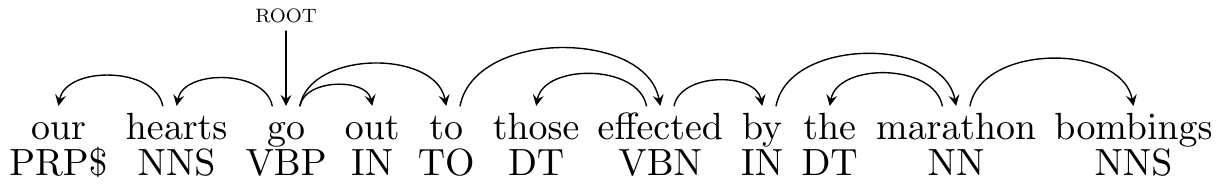}  
\caption{The dependency tree and the part of speech tags of a sample tweet.}
\label{fig:dependPlot1}
\end{figure}

\noindent
\newline
\emph{\textbf{Dependency Sub-trees: }}Much can be gained from the inclusion of sophisticated syntactic features such as dependency sub-trees in our speech act classifier. We used Kong et al.'s \cite{kong2014dependency} Twitter dependency parser for English (called the \emph{TweeboParser}) to generate dependency trees for our tweets. Dependency trees capture the relationship between words in a sentence. Each node in a dependency tree is a word with edges between words capturing the relationship between the words (a word either modifies or is modified by other words). In contrast to other syntactic trees such as \emph{constituency trees}, there is a one-to-one correspondence between words in a sentence and the nodes in the tree (so there are only as many nodes as there are words). Figure \ref{fig:dependPlot1} shows the dependency tree of an example tweet.

We extracted sub-trees of length one and two (the length refers to the number of edges) from each dependency tree.  Overall we collected 5,484 sub-trees that appeared at least five times. We then used a filtering process identical to the one used for n-grams, resulting in 1,655 sub-trees. There is a binary feature for each of these sub-trees indicating their appearance. 


\noindent
\newline
\emph{\textbf{Part-of-speech: }}Finally, we used the part-of-speech tags generated by the dependency tree parser to identify the use of adjectives and interjections (such as \emph{yikes}, \emph{dang}, etc). Interjections are mostly used to convey emotion and thus can signal expressions. Similarly adjectives can signal expressions or recommendations. We have two binary features indicating the usage of these two parts-of-speech.

\section{Supervised Speech Act Classifier}
We trained four different classifiers on our 3,313 binary features using the following methods: \emph{naive bayes (NB)}, \emph{decision tree (DT)}, \emph{logistic regression (LR)}, \emph{SVM}, and a baseline max classifier \emph{BL}. 
We trained classifiers across three granularities: \emph{Twitter-wide}, \emph{Type-specific}, and \emph{Topic-specific}. All of our classifiers are evaluated using 20-fold cross validation. 
Table \ref{table:all_f1} shows the performance of our five classifiers trained and evaluated on all of the data. We report the F1 score for each class. As shown in Table \ref{table:all_f1}, the logistic regression was the performing classifier with a weighted average F1 score of $.70$. Thus we picked logistic regression as our classier and the rest of the results reported will be for LR only. Table \ref{all:res} shows the average performance of the LR classifier for Twitter-wide, type and topic specific classifiers.



\begin{table}[!htbp]
\centering
\small
\begin{tabular}{@{}l|l|l|l|l|l|l|l@{}}
& As & Ex & Qu & Rc & Rq & Mis & Avg \\
\hline
BL & $0.$ & $.59$  & $0.$ & $0.$ & $0.$ & $0.$ & $.24$ \\
\hline
DT &  $.57$  &  $.68$  &  $.79$  &  $.32$  &  $0.$  &  $.29$  &  $.58$ \\
\hline
NB &  $.72$  &  $.76$  &  $.71$  &   \textbf{.40}  &  $0.$  &  $.41$  &  $.66$ \\
\hline
SVM & $.71$  &  $.80$  &  $.86$  &  $.35$  &  $.13$  &  $.43$  &  $.69$ \\
\hline
LR & \textbf{.73}  &  \textbf{.80}  &  \textbf{.87}  &  .30  &  \textbf{.16}  &  \textbf{.45}  & \textbf{.70}
\end{tabular}
\caption{F1 scores for each speech act category. The best scores for each category are highlighted.}

\label{table:all_f1}
\end{table}

\begin{table}[!htbp]
\centering
\small
\begin{tabular}{@{}l@{}|l|l|l|l|l|l|l@{}}
& As & Ex & Qu & Rc & Rq & Mis & Avg \\
\hline
Topic & $.73$ & $.87$ & $.98$ & $.57$ & $.03$ & $.26$ & \textbf{.74} \\
\hline
Type &  $.71$ & $.84$ & $.98$ & $.37$ & $.11$ & $.25$ & $.71$ \\
\hline
Twitter-wide  &  $.73$  &  $.80$  &  $.87$  &  $.30$  &  $.16$  &  $.45$  & $.70$ \\
\hline

\end{tabular}
\caption{F1 scores for Twitter-wide, type-specific and topic-specific classifiers.)}
\label{all:res}
\end{table}


The topic-specific classifiers' average performance was better than that of the type-specific classifiers ($.74$ and $.71$ respectively) which was in turn marginally better than the performance of the Twitter-wide classifier ($.70$). This confirms our earlier hypothesis that the more granular type and topic specific classifiers would be superior to a more general Twitter-wide classifier.  

 
Next, we wanted to measure the contributions of our semantic and syntactic features. To do so, we trained two versions of our Twitter-wide logistic regression classifier, one using only semantic features and the other using syntactic features. As shown in Table \ref{table:all_semsyn}, the semantic and syntactic classifiers' performance was fairly similar, both being on average significantly worse than the combined classifier. The combined classifier outperformed the semantic and syntactic classifiers on all other categories, which strongly suggests that both feature categories contribute to the classification of speech acts.

\begin{table}[!htbp]
\centering
\begin{tabular}{@{}l|l|l|l|l|l|l|l@{}}
& As & Ex & Qu & Rc & Rq & Mis & Avg \\
\hline

Sem  &  .71  &  .80  &  .62  &  .22  &  0.  &  .23  &  .64 \\

Syn  &  .59  &  \textbf{.81}  &  \textbf{.94}  &  .12  &  0.  &  0.  &  .62 \\
\hline
All & \textbf{.73}  & .80  &  .87  & \textbf{ .30}  &  \textbf{.16}  &  \textbf{.45}  & \textbf{.70}
\end{tabular}
\caption{F1 scores for each speech act category for semantic and syntactic features.}

\label{table:all_semsyn}
\end{table}


Finally, we compared the performance of our classifier (called TweetAct) to a logistic regression classifier trained on features proposed by, as far as we know, the only other supervised Twitter speech act classifier by Zhang et al. (called Zhang). Table \ref{table:compare} shows the results. Not only did our classifier outperform the Zhang classifier for every class, both the semantic and syntactic classifiers (see Table \ref{table:all_semsyn}) also generally outperformed the Zhang classifier.

\begin{table}[!htbp]
\centering
\small
\begin{tabular}{@{}l|l|l|l|l|l|l|l@{}}
& As & Ex & Qu & Rc & Rq & Mis & Avg \\
\hline

Zhang  &  $.67$  &  $.60$ &  $.73$  &  $.18$  &  $0.$  &  $.19$  &  $.59$ \\
TweetAct & \textbf{.73}  & \textbf{.80}  &  \textbf{.87}  & \textbf{ .30}  &  \textbf{.16}  &  \textbf{.45}  & \textbf{.70}
\end{tabular}
\caption{F1 scores for each speech act category for our classifier compared to the Zhang classifier.}

\label{table:compare}
\end{table}



\section{Conclusions and Future Work}
In this paper, we presented a supervised speech act classifier for Twitter. We treated speech act classification on Twitter as a multi-class classification problem and came up with a taxonomy of speech acts on Twitter with six distinct classes. We then proposed a set of semantic and syntactic features for supervised Twitter speech act classification. Using these features we were able to achieve state-of-the-art performance for Twitter speech act classification, with an average F1 score of $.70$. Speech act classification has many applications; for instance we have used our classifier to detect rumors on Twitter in a companion paper \cite{vosoughi_rd_2016}.




\small
\bibliographystyle{aaai}
\bibliography{sac}

\end{document}